%% file: egpaper_for_review.tex
\documentclass[10pt,twocolumn,letterpaper]{article}

\usepackage{wacv}
\usepackage{times}
\usepackage{epsfig}
\usepackage{graphicx}
\usepackage{amsmath}
\usepackage{amssymb}
\usepackage{multirow}
\usepackage{footnote}
\usepackage{caption}

\usepackage[pagebackref=true,breaklinks=true,letterpaper=true,colorlinks,bookmarks=false]{hyperref}

\wacvfinalcopy 

\def\wacvPaperID{37} 

\ifwacvfinal\fi

\DeclareMathOperator*{\argmax}{arg\,max}
\begin{document}

\title{Proposal-free Temporal Moment Localization of a Natural-Language Query in Video using Guided Attention}

\author{Cristian Rodriguez-Opazo$^{1,2}$ \qquad Edison Marrese-Taylor$^{3}$ \qquad Fatemeh Sadat Saleh$^{1,2}$\\ \qquad Hongdong Li$^{1,2}$ \qquad Stephen Gould$^{1,2}$ \\
\\
${}^{1}$Australian National University, 
${}^{2}$Australian Centre for Robotic Vision (ACRV)\\{\tt\small \{cristian.rodriguez, fatemehsadat.saleh, hongdong.li, stephen.gould\}@anu.edu.au}\\
${}^{3}$Graduate School of Engineering, The University of Tokyo\\
{\tt\small emarrese@weblab.t-utokyo.ac.jp}\\
}

\maketitle
\ifwacvfinal\thispagestyle{empty}\fi


\input{introduction}

\input{related_work}

\input{method}
\input{experiments}
\input{conclusion}
\section{Acknowledgement}

This research is supported in part by the Australia Research Council Centre of Excellence for Robotics Vision (CE140100016). We gratefully acknowledge the GPU gift donated by NVIDIA Corporation. We thank all anonymous reviewers for their constructive comments.

{\small
\bibliographystyle{ieee}
\bibliography{egbib}
}

\end{document}

%% file: introduction.tex
\begin{abstract}
This paper studies the problem of temporal moment localization in a long untrimmed video using natural language as the query. Given an untrimmed video and a query sentence, the goal is to determine the start and end of the relevant visual moment in the video that corresponds to the query sentence.  While most previous works have tackled this by a propose-and-rank approach, we introduce a more efficient, end-to-end trainable, and proposal-free approach that is built upon three key components: a dynamic filter which adaptively transfers language information to visual domain attention map, a new loss function to guide the model to attend the most relevant part of the video, and soft labels to cope with annotation uncertainties.  Our method is evaluated on three standard benchmark datasets, Charades-STA, TACoS and ActivityNet-Captions. Experimental results show our method outperforms state-of-the-art methods on these datasets, confirming the effectiveness of the method.  We believe the proposed dynamic filter-based guided attention mechanism will prove valuable for other vision and language tasks as well.  
\end{abstract}



\section{Introduction}



Vision-and-language understanding is an important problem in computer vision, drawing increasing attention from the community over the past a few years, motivated by its vast potential applications. This setting includes tasks such as video captioning and video question answering. While promising results have been achieved in these tasks, a fundamental issue remains to be tackled, namely, that these informative video segments need to be manually trimmed (pre-segmented) and often aligned with the relevant textual descriptions that accompany them. 

Since searching for a specific visual event over a long video sequence is difficult and inefficient to do manually, even for a small number of videos, automated search engines are needed to deal with this requirement, especially when dealing with a massive amount of video data.  It is clear that these search engines have to retrieve videos not only based on the video metadata, but they also must exploit their internal information in order to accurately localize the required information/segment. 

\begin{figure}[t]
 \centering
 \includegraphics[width=0.471\textwidth]{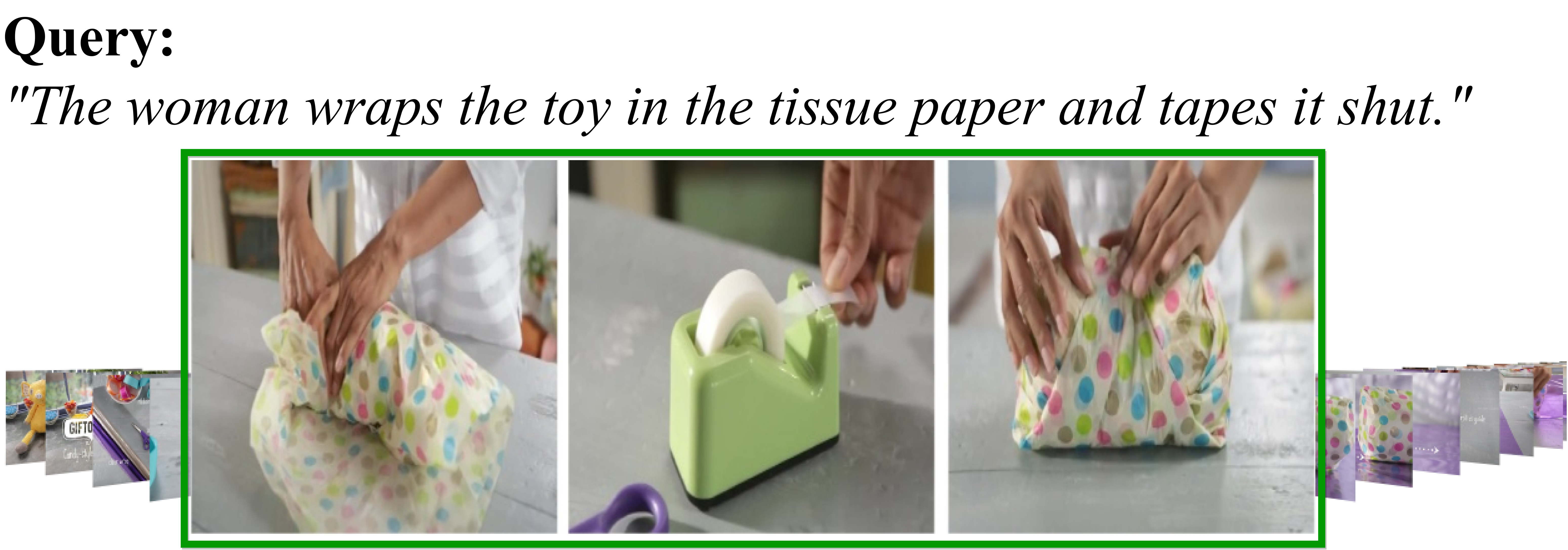}
 \caption{An illustration of temporal localization of a natural language query in an untrimmed video. Given a query and a video the task is temporally localize the starting and ending of the sentence in the video.}
 \label{fig:example}
    \vspace{-6mm}
\end{figure}

In light of this, automatically recognizing \textit{when} an activity is happening in a video has become a crucial task in computer vision. Its applicability to other research areas such as video surveillance and robotics \cite{Liu_ICRA_2018}, among others, has also helped bring interest into this task. Earlier works in this area focused on {\em temporal action localization} \cite{richard2018neuralnetwork, Lin2017a, Xu, Zhao, Escorcia2016, Chao2018, GaoYSCN17}, which attempted to localize ``interesting'' actions in a video from a predefined set of actions. However, this approach constrains the search engine to a relatively small and unrealistic set of queries from users. 

To address this issue the task of ``temporal action localization with natural language'' has been proposed recently \cite{Gao_2017_ICCV,Hendricks_2017_ICCV}. Given a query, the goal is to determine the start and end locations of the associated video segment in a long untrimmed video. In this context, we are specifically interested in the problem of natural-language-based temporal localization, or temporal sentence localization in the video. Formally, given an untrimmed video and a query in natural language, the task is to identify the start and end points of the video segment in response to it, therefore effectively locating the temporal segment (i.e., moment) that best corresponds to the given query, as depicted in Figure \ref{fig:example}. 

Current approaches to the localization problem in computer vision, either spatial or temporal, mainly focus on creating a good multi-modal embedding space and generating proposals based on the given query. In these \textit{propose and rank} approaches, candidate regions are first generated by a separate method and then fed to a classifier to get the probabilities of containing target classes, effectively ranking them. Despite the relative success of these approaches, this setting is ultimately restrictive in scope since it uses predefined clips as candidates, making it hard to extend for videos with considerable variance in length. 



To this end, we propose an approach that does not rely on candidate generation or ranking, being able to directly predict the start and end times given a query in natural language. Our model is guided by a dynamic filter, which is responsible for matching the text and video feature embeddings, and a principal attention mechanism which encourages the model to focus on the features within of segment of interest. To the best of our knowledge, our approach is the first to do so\footnote{Code and features can be found in \url{https://github.com/crodriguezo/TMLGA}}.

Recent works on temporal action localization with natural language \cite{ghosh2019excl} has adopted an approach akin to machine reading comprehension (MC) \cite{chen_reading_2017}, but in a multi-modal setting. Similar to ours, these models are trained in an end-to-end manner. Specifically, they maximize the likelihood of correctly predicting the start and end frames associated with a given query, analogous to predicting the text span corresponding to the correct answer in MC. We note, however, that annotating the start and end of a given activity in a video is highly subjective, as evidenced by relatively low inter-annotator agreement \cite{sigurdsson_ICCV_2017,Alwassel_2018_ECCV,ferrari_diagnosing_2018}. In light of this, our model incorporates annotation subjectivity in a simple yet efficient manner, obtaining increased performance.

We conduct experiments on three challenging datasets, Charades-STA \cite{Gao_2017_ICCV}, TACoS  \cite{tacos} and Activity Net Captions \cite{Krishna_2017_ICCV}, demonstrating the effectiveness of our proposed method and obtaining state-of-the-art performance on them. Our results also empirically demonstrate the effectiveness of our attention-based guidance mechanism, and of incorporating the subjective nature of the annotations into the model, ultimately validating our proposed approach through ablation analysis.



%% file: related_work.tex
\section{Related Work}

\subsection{Temporal Action Localization}

The task of temporal action localization aims to solve the problem of recognizing and determining temporal boundaries of action instances in videos. Since activities (in the wild) consist of a diverse combination of actors, actions and objects over various periods of time, earlier work focused on classification of video clips that contained a single activity, i.e., where the videos were trimmed. 

More recently there has also been significant work in localizing activities in longer, untrimmed videos. For example, Shou et al. \cite{Shou_2016_CVPR} trained C3D \cite{tran2015learning} with a localization loss and achieved state-of-the-art performance on THUMOS \cite{idrees_thumos_2017}. On the other hand, Ma et al. \cite{ma_learning_2016} used a temporal LSTM to generate frame-wise prediction scores and then merged the detection intervals based on the predictions. Singh et al. \cite{singh_multi-stream_2016} extended the two-stream \cite{simonyan2014two} framework with person detection and bi-directional LSTMs and achieved state-of-the-art performance on the MPII-Cooking dataset \cite{rohrbach_dataset_2015}. 

Escorcia et al. \cite{escorcia_daps:_2016} took a different approach and introduced an algorithm for generating temporal action proposals from long videos, which are used to retrieve temporal segments that are likely to contain actions. Lin et al. \cite{lin_single_2017} proposed an approach based on 1D temporal convolutional layers to skip the proposal generation step via directly detecting action instances in untrimmed video.

The major limitation of these action localization methods is that they are restricted to a pre-defined list of actions. As it is non-trivial to design a label space which has enough coverage for such activities without losing useful details in users' queries this approach makes it difficult to cover complex activity queries. 

\subsection{Temporal language-driven moment localization}

Language-driven temporal moment localization is the task of determining the start and end time of the temporal video segment that best corresponds to a given natural language query. Essentially, this means to use natural language queries to localize activities in untrimmed videos. While the language-based setting allows for having an open set of activities, it also corresponds to a more natural query specification, as it directly includes objects and their properties as well as relations between the involved entities. 

The work of Hendricks et al. \cite{Hendricks_2017_ICCV} and Gao et al. \cite{Gao_2017_ICCV} are generally regarded as pioneer on this task. While Hendricks et al. \cite{Hendricks_2017_ICCV} proposed to learn a shared embedding for both video temporal context features and natural language queries, suitable for matching candidate video clips and language queries using a ranking loss and handcrafted heuristics, Gao et al. \cite{Gao_2017_ICCV} proposed to generate candidate clips using temporal sliding windows which are later ranked based on alignment or regression learning objectives. 

The research line defined by Gao et al. \cite{Gao_2017_ICCV}, where proposals are generated using temporal sliding windows was later extended by Ge et al. \cite{ge2019mac}, which leverage activity classifiers to help encode visual concepts, and add an \textit{actionness score} to help capture the semantics from verb-object pairs in the queries. Recently, Liu et al. \cite{liu2018attentive} also resorted to sliding windows for generating proposals, but used a memory attention model when matching each proposal to the input query. Despite their simplicity and ability to provide coarse control over the frames that are evaluated, the main problem with these methods is that the matching mechanism between the candidate proposals and the query is computationally expensive. 
 

To tackle this issue some approaches have focused on reducing the number of temporal proposals generated. These methods generally focus on producing query-guided or query-dependent video clip proposals, skipping unlikely clips from the matching step and thus speeding up the whole process. In this context, Chen et al. \cite{chen-etal-2018-temporally} propose to capture frame-by-word interactions between video and language and then score a set of temporal candidates at multiple scales to localize the video segment that corresponds to the query. Similarly, Xu et al. \cite{xu2019multilevel} propose a multilevel model to tightly integrate language and vision features and then use a fine-grained similarity measure for query-proposal matching.

A slightly different but related approach is proposed by Hendricks et al. \cite{Hendricks_2018_EMNLP}, where the video context is modeled as a latent variable to reason about the temporal relationships. The work of Zhang et al. \cite{zhang2018man} further improved on this by utilizing a graph structured network to model temporal relationships among different moments, therefore addressing semantic and structural misalignment problems. On the other hand, Chen et al. \cite{sap2019} focused on the proposal generation step, integrating the semantic information of the natural language query into the proposal generation process to get discriminative activity proposals. Although previous methods use techniques to directly generate candidate moment representations aligned with language semantics instead of fetching video clips independently, they still depend on ranking a fixed number of temporal candidates in each video, leading to inefficiencies. 


More recently, methods that go beyond the \textit{scan and localize} approach, which can therefore directly output the temporal coordinates of the localized video segment have been proposed. For example, Yuan et al. \cite{yuan2018find} used a co-attention based model for temporal sentence localization. In this context, some models also resort to reinforcement learning to dynamically observe a sequence of video frames conditioned on the given language query. Concretely, Wang et al. \cite{wang2019language} train a recurrent neural network for language-driven temporal activity localization using this approach, while also utilizing Faster R-CNN~\cite{ren_NIPS_2015_faster} trained on the Visual Genome dataset~\cite{krishnavisualgenome} to obtain regional visual features and incorporate more semantic concepts to the model. Similarly, Hahn et al. \cite{hahn2019tripping} use this approach and learn how to skip around the video, therefore being able to more easily localize relevant 
clips in long videos. Instead of simply concatenating the video representation and query embedding, their approach uses a gated attention architecture to model textual and visual representations in order to align the text and video content. 

Finally, Ghosh et al. \cite{ghosh2019excl} proposes a simpler approach that does not rely on reinforcement learning and does not either involve retrieve and re-ranking multiple proposal segments. Their approach focuses on predicting the start and end frames by leveraging cross-modal interactions between the text and video. In this context, our method proposes a simple yet effective proposal-free approach which makes it more practical to use.

%% file: method.tex
\section{Proposed Approach}

\begin{figure*}[!ht]
    \centering
    \includegraphics[width=0.80\textwidth]{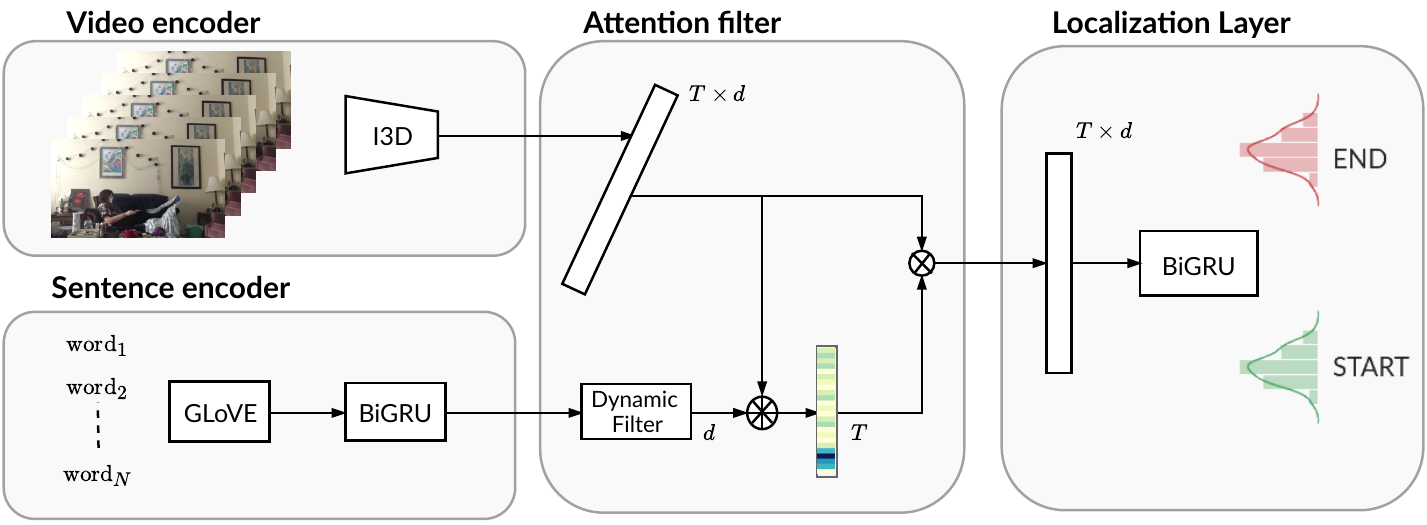}
    \caption{Overview of our method with its four modules: sentence and video encoders to extract features from each modality; a dynamic filter to transfer language information to video, and a localization layer to the starting and ending points.}
    \label{fig:overview}
    \vspace{-5mm}
\end{figure*}

Let $V \in \mathcal{V}$ be a video that can be characterized as a sequence of frames such that $V = \{v_t\}$ with $t = 1, \ldots, l$. Each video in $\mathcal{V}$ is annotated with a natural language passage $S \in \mathcal{S}$ where $S$ is a sequence of words $S = \{s_j\}$ with $j = 1, \ldots, m$, which describes what is happening in a certain period of time. Formally, this interval is defined by $t^s$ and $t^e$, the starting and ending points of the annotations in time, respectively. 

We propose a model that is trained end-to-end on a set of example tuples of annotated videos $(V_k, S_k, t_k^s, t_k^e)$. Although in the data a given video may be annotated with more than one single moment, and one natural language description may be associated to multiple moments, in this work we assume each derived case as an independent, separate training example. Given a new video and sentence tuple $(V_r, S_r)$, our model predicts the most likely temporal localization of the contents of $S_r$ in terms of its start and end positions $t_r^{s\star}$ and $t_{r}^{e\star}$ in the video, therefore effectively solving the problem of temporal localization of sentences in videos. In the following, for simplicity we drop the index $k$ associated to each training example. 

Our model is designed in a modular way, offering more flexibility over previous work. There are four main components which we proceed to describe in the following sections. 
Figure \ref{fig:overview} shows an overview of our proposed approach.

\subsection{Video Encoder}
\label{sec:method:video}


As discussed earlier, previous works on temporal sentence localization in videos mostly rely on proposal generation, either using sliding windows or other heuristics \cite{Gao_2017_ICCV, Hendricks_2017_ICCV, ge2019mac, liu2018attentive}. The process of producing many temporal segment candidates is computationally expensive, even though its efficiency can be improved if the proposals are processed in parallel. Moreover, proposal-based mechanisms neglect time dependencies across segments, treating them independently thus ultimately failing to effectively capture the temporal information in the input video. 



Inspired by recent works in one-shot object detection \cite{redmon2016you, liu2016ssd} , we propose a video encoding layer that generates a visual representation summarizing spatio-temporal patterns directly from the raw input frames. Concretely, given an input video $V$, let $F_V(V)$ be our video encoding function mapping the $l$ input video frames to a sequence of vectors $G =\{g_i \in \mathbb{R}^{d_v}\}$, $i=1, \ldots, n$, with features that capture high-level visual semantics in the video. Note that the length of the input vector in frames $l$ and the number of output features $n$ may differ, which is why we denote them differently. 

Because of the encoding of the video, the location of the annotated natural language description needs to be re-scaled to match the new feature-wise setting. We apply the mapping $\tau = (t \cdot n \cdot \text{fps}) / l$ to convert from frame/feature index to time. Concretely, $t^s$ and $t^e$ are converted into $\tau^s$ and $\tau^e$ corresponding to specific integer feature positions such that $\tau^s, \tau^e \in [1, \ldots, n]$.


Specifically, in this work we model $F_V$ using I3D \cite{carreira2017quo}. This method inflates the 2D filters of a well-known network e.g., Inception \cite{szegedy2015going, ioffe2015batch} or ResNet \cite{He_2016_CVPR} for image classification to obtain 3D filters, helping us better exploit the spatio-temporal nature of video. However, note that our video encoder later is generic, so other alternatives such as C3D \cite{tran2015learning} could be utilized instead.

\subsection{Sentence Encoder}
\label{sec:method:sentence}


The language encoder aims at generating a semantically rich representation of the natural language query that is useful for localizing relevant moments in the video. We model our encoder as a function $F_S(S)$ that maps each word $s_j$ for $j = 1, \ldots, m$ to a semantic embedding vector $h_j \in \mathbb{R}^{d_s}$, where $d_s$ defines the hidden dimension of the obtained sentence representation. 

Although our sentence encoding module is generic, in this work we rely on a bi-directional GRU \cite{chung2014empirical} on top of pre-trained word embeddings. Specifically, we make use of GloVe \cite{pennington_glove:_2014}, which are vectors pre-trained in a large collection of text documents. In this setting, our query encoding function $F_S$ is parametrized by both the GloVe embeddings and the GRU. Finally, to obtain a fixed-length sentence representation we utilize a mean pooling layer over the hidden states obtained from the GRU, obtaining a final summarized query representation $\bar h$.

\subsection{Guided Attention}
\label{sec:method:filter}

After encoding both the input sentence and video we utilize an attention-based {\em dynamic filter} \cite{jia2016dynamic, li2017tracking, Gavrilyuk_2018_CVPR, zhang2018man}. The motivation behind this is to allow the model to generate filters to be applied over the video features that dynamically change given the sentence query, effectively reacting to specific parts of the video embedding and thus providing the model with clues about the location. 


Concretely, we first reduce the dimensionality of the sentence embedding $d^{s}$ and the video embedding $d^{v}$ to the same space of size $d$ using a fully connected network, and apply a filter function $\theta$ as follows.
\begin{equation} 
    \label{eq:attention}
    \theta(x) = \text{tanh}(W_{\theta} x + b_{\theta}) \in \mathbb{R}^{d}
\end{equation}

As seen in Equation \ref{eq:attention}, our filter function $\theta(\cdot)$ is a single-layer fully-connected neural network. The sentence representation $\bar h$ is fed into our function and the obtained filter is later used to create a temporal attention over the video features $G$. Specifically, we apply a softmax over the inner-product between each video feature $g_i$ and the output of the filter $\theta(\bar h)$, as follows,
\begin{align}
    A &= \text{softmax} \left( \cfrac{G^{\intercal} \theta(\bar{h})}{\sqrt{n}} \right) \in \mathbb{R}^{n} \\
    \bar{G} &= A \odot G \in \mathbb{R}^{n \times d}
\end{align}
where $\odot$ denotes the Hadamard product, and the $1/\sqrt{n}$ constant is used to re-scale the product for better training stability \cite{vaswani2017attention}. As a result of these operations, each video feature is scaled by the attention filter based on the natural language query.

Given a category of semantically similar natural language queries, for example describing the same type of action, we would like our model to focus on the spatio-temporal features that most likely describe and generalize these semantics across all examples where they are relevant, regardless of the additional context in the videos. We therefore argue that the most relevant features should fall inside the time boundary ($\tau_s$ to $\tau_e$) defined by the starting and ending points of the target locations to be predicted. Although features from outside this segment could also contain useful information for the localization task, we hypothesize that by exploiting these features the model should attain less generalization power, as these features are not likely to capture patterns that appear in the majority of different videos containing a given type of action.


In light of this, we encourage our model to attend these relevant features and therefore improve its generalization capabilities. Concretely, we guide our attention mechanism to put focus on these features using a loss function on the output, as follows.
\begin{equation}
    L_{att} = - \sum_{i=1}^n (1- \delta_{\tau^s \leq i \leq \tau^e}) \log(1-a_i) 
    \label{eq:att_focus}
\end{equation}
where $\delta$ is the Kronecker delta 
and $a_i$ is the $i$th column in the attention matrix $A$.

\subsection{Localization Layer}
\label{sec:method:localization}


The localization layer is in charge of predicting the starting and ending points of the moment in the video, using the previously constructed sequence of attended video features $\bar{g_i} \text{ for } i=1, \ldots, n$. 

Humans have difficulty agreeing on the starting and ending time of an action inside a video, as evidenced by the low inter-annotation agreement in the relevant datasets for temporal localization  \cite{sigurdsson2017actions, Alwassel_2018_ECCV}. Considering that this is therefore a highly subjective task, we take a probabilistic approach and propose to use {\em soft-labels} \cite{salimans2016improved,szegedy_rethinking_2016} to model the uncertainty associated to the labels. 

The localization layer starts by further contextualizing the attended video features $\bar{g_i}$ utilizing a 2-layer bidirectional GRU \cite{chung2014empirical}. Then, we utilize two different fully connected layers to produce scores associated to the probabilities of each position being the start/end of the location. For each case, we take the softmax of these scores and thus obtain vectors $ \hat{\boldsymbol {\tau}}^s, \hat{\boldsymbol{\tau}}^e \in \mathbb{R}^{n}$ containing a categorical probability distribution for the predicted start and end positions, respectively. 

To model annotation uncertainty, we take $\tau^s$ and $\tau^e$ and create two target categorical distribution vectors $\boldsymbol{\tau}^s \sim \mathcal{N}(\tau^s$, $1) \in \mathbb{R}^{n}$ and $ \boldsymbol{\tau}^e \sim \mathcal{N}(\tau^e$, $1) \in \mathbb{R}^{n}$ respectively, where $\mathcal{N(\mu, \sigma)}$ denotes a quantized Gaussian distribution centered at $\mu$, with standard deviation $\sigma$ ---discretizing a Gaussian distribution over the interval $[1, n]$. We train our model to minimize the Kullback-Leibler divergence between the predicted and ground truth probability distributions, as follows.
\begin{equation}
    L_{KL} = \displaystyle D_{\text{KL}}(\hat{\boldsymbol{\tau}}^s \parallel \boldsymbol {\tau}^s) + \displaystyle D_{\text{KL}}(\hat{\boldsymbol {\tau}}^e \parallel \boldsymbol {\tau}^e)
    \label{eq:kl_div}
\end{equation}
where $D_{\text{KL}}$ is the Kullback-Leibler divergence. The final loss for training our method is the sum of the two individual losses defined previously.
\begin{equation}
    Loss = L_{KL} + L_{att}
\end{equation}

During inference, we predict the starting and ending positions using the most likely locations given by the estimated distributions:
\begin{equation}
    \hat{\tau}^s = \argmax(\hat{\boldsymbol{\tau}}^s) \quad \hat{\tau}^e = \argmax(\hat{\boldsymbol{\tau}}^e)
\end{equation}
These values correspond to positions in the feature domain of the video, so we convert them back to time positions as explained previously.

%% file: experiments.tex
\section{Experiments}

In this section, we first describe the datasets used in our experiments and give some details about our learning procedure. Then, we provide an ablation study on the effect of different components and we compare our approach with other methods. Finally, we provide a qualitative visualization of the predicted localization and attention.


\subsection{Datasets}
To evaluate our proposed approach we work with three challenging datasets for temporal natural language-driven moment localization, Charades-STA \cite{Gao_2017_ICCV}, TACoS \cite{tacos} and ActivityNet Caption \cite{caba2015activitynet, Krishna_2017_ICCV}, which are widely utilized in previous works for evaluating models on our task. 
    
\noindent
\textbf{Charades-STA}: built upon the Charades dataset \cite{sigurdsson2016hollywood} which provides time-based annotations using a pre-defined set of activity classes, and general video descriptions. In Gao et al. \cite{Gao_2017_ICCV}, the sentences describing the video are semi-automatically decomposed into smaller chunks and aligned with the activity classes, which are later verified by human annotators. As a result of this process, the original class-based activity annotations are effectively associated to their natural language descriptions, totalling 13,898 pairs. We use the predefined train and test splits, containing 12,408 and 3,720 moment-query pairs respectively. Videos are 31 seconds long on average, with 2.4 moments on average, each being 8.2 seconds long on average.

\noindent
\textbf{MPII TACoS} \cite{tacos} has been built on top of the MPII Compositive dataset \cite{rohrbach2012script}. It consists of detailed temporally  aligned  text  descriptions  of  cooking  activities. The average length of videos is 5 minutes. A significant challenge in TACoS dataset is that descriptions span over only a few seconds because of the atomic nature of queries such as ‘takes out the knife’ and ‘chops the onion’ (8.4\% of them are less than 1.6s long). Such short queries allow a smaller margin of error. In total, there are are 18,818 pairs of sentence and video clips. We use the same split as in \cite{Gao_2017_ICCV}, consisting of 50\% for training, 25\% for validation and 25\% for testing.

\noindent
\textbf{ActivityNet Caption (ANet-Cap)}: a large dataset built on top of ActivityNet \cite{caba2015activitynet}, which contains data derived from YouTube and annotated for the tasks of activity recognition, segmentation and prediction. ANet-Cap further improves the annotations in ANet by incorporating descriptions for each temporal segment in the videos, totalling up to 100K temporal descriptions annotations over 20K videos. These have an average length of 2.5 minutes and are associated to over 200 activity classes, making the content much more diverse compared to Charades-STA. The temporally annotated moments are 36 seconds long on average, with videos containing 3.5 moments on average. Besides moments being longer than in Charades-STA, we find that their associated natural language descriptions are also longer, besides using a more varied and richer vocabulary. We utilize the predefined train and validation splits in our experiments. Unlike Charades, Activity-Net contains a moment covering the entire video.
\subsection{Implementation Details}
Pre-processing for the natural language input in the case of Charades-STA is minimal, as we simply tokenize and keep all the words in the training data. In the case of ANEt-Cap, we pre-process the text with spacy\footnote{\url{https://spacy.io}} and replace all named entities as well as proper nouns with special markers. Finally, we truncate all sentences to a maximum length of 30 words and replace all tokens with frequency lower than 5 in the corpus with a special \textit{UNK} marker. The language encoder uses a hidden state of size $256$, without fine-tuning the pre-trained GloVe embeddings. 

When it comes to the video encoder, we first pre-process the videos by extracting features of size $1024$ using I3D with average pooling, taking as input the raw frames of dimension $256 \times 256$, at $25$ fps. For Charades-STA, we use the pre-trained model released by \cite{carreira2017quo} trained on Charades. For Anet-Cap we use the model pre-trained on the kinetics400 dataset \cite{Kinetics400} released by the same authors, which we also fine-tune on ANet-Cap afterwards. 

All of our models are trained in an end-to-end fashion using Adam \cite{kingma_adam} with a learning rate of $10^{-4}$ and weight decay $10^{-3}$. To prevent over-fitting, we add dropout $0.5$ between the two layers in the localization module, which has a hidden size of $256$. In addition to this, we also apply a simple data augmentation technique that consists on creating new examples by randomly cropping segments out from the initial part of the videos. We do this whenever the random cropping does not overlap with the locations of the annotations.

\subsection{Evaluation Metric} 
We evaluate our model by computing the temporal Intersection over Union (tIoU) at different thresholds, which we denote using the $\alpha$ parameter. In this setting, for a given value of $\alpha$, whenever a given predicted time window has an intersection with the gold-standard that is above the $\alpha$ threshold, we consider the output of the model as correct. Following previous work, we also report the mean tIoU (mIoU) on the ANet-Cap dataset, helping make our comparisons more comprehensive.

\begin{table}[t]
    \centering
    \scalebox{0.9}{
    \begin{tabular}{|l|c|c|c|}
         \textbf{Method} & $\alpha=0.3$ & $\alpha=0.5$  & $\alpha=0.7$ \\ \hline \hline
         
        NLL & 60.91 & 43.66 & 27.07 \\
        KL & 66.69 & 47.20  & 29.35 \\
        NLL + AL &  66.64 & 47.53 &  29.89\\
        KL + AL & \textbf{67.53} & \textbf{52.02} & \textbf{33.74} \\  \hline
    \end{tabular}
    }
    \caption{Ablation study on the impact of the guided attention and soft-labeling on Charades-STA.}
    \label{tab:ablation:res}
    \vspace{-4mm}
\end{table}


\subsection{Ablation Study}
To show the effectiveness of some introduced components, we perform several ablation studies on the Charades-STA dataset. Concretely, we focus on the soft-labeling technique and the usage of the attention loss $L_{att}$. For the latter we simply experiment omitting the term for the calculation of the gradients. For the former, we replace the $L_{KL}$ loss with a likelihood-based loss similar to \cite{ghosh2019excl}, as follows:
\begin{equation}
    L_{NLL} = - \log(\boldsymbol {\hat{\tau}}^s[\tau^s]) - \log(\boldsymbol {\hat{\tau}}^e[\tau^e]) 
    \label{eq:nll}
\end{equation}
where $\boldsymbol{\hat{\tau}}^s$ and $\boldsymbol{\hat{\tau}}^e$ are the predicted probability distributions and $\tau^s$ and $\tau^e$ are the respective indices from the ground-truth annotations.

We first compare our {\em soft-labeling} approach with the previously mentioned likelihood-based loss (NLL). As shown in Table \ref{tab:ablation:res}, modeling the subjectivity of the labeling process using soft-labels and our distribution-matching loss (KL) leads to a significant improvement in the performance of our method, both in terms of retrieving and localizing the full extent of the queries in the given videos.


We also evaluate the contribution of the attention loss $L_{att}$ to our pipeline. According to the results in Table \ref{tab:ablation:res}, adding the attention loss (AL) leads to a consistent improvement in the performance of our method, both when modeling soft-labels and when not. This confirms our hypothesis that the most generalizable features are likely to be located within the boundaries of the query segment in the video. Finally, the synergy of our two proposed techniques can be seen in the last row of Table \ref{tab:ablation:res}.




\subsection{Comparison to the State-of-the-Art}
We compare the performance of our proposed approach on both datasets against several prior work baselines. 
\textbf{Proposal-based methods}: We compare our approach to a broad selection of models based on proposal generation, including MCN \cite{Hendricks_2017_ICCV}, TGN \cite{chen-etal-2018-temporally}, MAN \cite{zhang2018man}, as well as some recent work such as SAP \cite{sap2019}, MLVI \cite{xu2019multilevel} and ACRN \cite{liu2018attentive}.

\noindent
\textbf{Reinforcement-learning-based methods}: We compare our results to TripNet \cite{hahn2019tripping} and SMRL \cite{wang2019language}, both of which utilize RL to learn how to jump through the video until the correct localization is found.

\noindent
\textbf{Proposal-free methods}: We consider two recent works, ABLR \cite{yuan2018find} and ExCL \cite{ghosh2019excl}, both aiming for proposal-free moment localization. Similar to ours, these techniques utilize the complete video representation to predict the start and end of a relevant segment. However, our approach is different since it models the uncertainty of the labeling process. Note also that while ABLR utilizes a co-attention layer, ExCL does not rely on attention layers at all.


Comparing the performance of our method in the \textbf{Charades-STA} benchmark, our proposed approach outperforms all the baselines by a large margin, as can be seen in Table \ref{tab:csta:res}. Its mean temporal intersection over union is $48.22$ reflecting the capability of our method to correctly identify the correct temporal extent of the natural language query. As can also be seen in the performance at $\alpha=0.7$ and $\alpha=0.9$ where our method obtains $33.74$ and $9.68$ accuracy for those thresholds.

\textbf{TACoS} is a challenging benchmark not only because the length of the videos is much longer than Charades-STA, but also because it presents a bigger variability of segment duration for a query. Since we are not processing videos using proposals these types of videos, our localization layer could have difficulties predicting the precise spans of queries. Despite that our method outperforms all previous methods at $\alpha = 0.7$, showing the robustness of our approach.

\textbf{ANet-Cap} is another challenging dataset similar to TACoS with a significant variability of the duration of the segments. However, as shown in Table \ref{tab:anetcap:res}, our method yields good performance at different levels of tIoU. In particular, it outperforms all previous methods at $\alpha$ 0.1 and 0.7, showing the effectiveness of our method to recall the correct temporal extent of the sentence query. Although our method cannot outperform the performance of ABLR at $\alpha$ 0.3 and 0.5, it yields better mIoU than previous methods in this benchmark, as can be seen in Table \ref{tab:anet:miou}. It is important to note that in this case we do not compare with ExCL \cite{ghosh2019excl} since their reported results have more than 3,300 missing videos. 


\begin{table}[t]
    \centering
    \scalebox{0.9}{
    \begin{tabular}{|l|c|c|c|}
         \textbf{Method} & $\alpha=0.3$ & $\alpha=0.5$  & $\alpha=0.7$ \\ \hline
          \hline
         Random & - & 8.51 & 3.03 \\ 
         CTRL \cite{Gao_2017_ICCV} & - & 21.42 &  7.15 \\
         ABLR \cite{sap2019} & - & 24.36 & 9.01 \\
         SMRL\cite{wang2019language} & - & 24.36 & 11.17 \\
         SAP \cite{sap2019} & - & 27.42 & 13.36\\
         MLVI \cite{xu2019multilevel} & 54.70 & 35.60 & 15.80 \\
         TripNet \cite{hahn2019tripping} & 51.33 & 36.61 & 14.50 \\
         ExCL \cite{ghosh2019excl} & 65.10 & 44.10 & 23.30 \\
         MAN \cite{zhang2018man} & - & 46.53 & 22.72 \\  \hline
         Ours & \textbf{67.53} & \textbf{52.02} & \textbf{33.74}  \\  \hline
    \end{tabular}
    }
    \caption{Accuracy on Charades-STA for different tIoU $\alpha$ levels. Results for ABLR are as reported by \cite{sap2019}.}
    \label{tab:csta:res}
    \vspace{-1mm}
\end{table}

\begin{table}[t]
    \centering
    \scalebox{0.9}{
    \begin{tabular}{|l|c|c|c|}
         \textbf{Method} & $\alpha=0.3$ & $\alpha=0.5$  & $\alpha=0.7$ \\ \hline
         \hline
         
        MCN  \cite{Hendricks_2017_ICCV} & 1.64  & 1.25  & - \\
        ABLR \cite{yuan2018find} & 18.90 & 9.30 & - \\ 
        CTRL \cite{Gao_2017_ICCV} & 18.32 & 13.30 &  - \\
        ACRN \cite{yuan2018find} & 19.52 & 14.62 &  - \\
        TGN \cite{chen-etal-2018-temporally} & 21.77 & 18.90 & -\\ 
        TripNet  \cite{hahn2019tripping} & 23.95 & 19.17 & 9.52 \\ 
ExCL \cite{ghosh2019excl} & \textbf{44.20} & \textbf{28.00} & 14.60 \\ \hline
Ours & 24.54 & 21.65 & \textbf{16.46} \\ \hline
    \end{tabular}
    }
    \caption{Accuracy on TACoS for different intersection over union $\alpha$ levels.}
    \label{tab:tacos:res}%
    \vspace{-1mm}
\end{table}

\begin{table}[t]
    \centering
    \scalebox{0.9}{
    \begin{tabular}{|l|c|c|c|c||c|}
         \textbf{Method} & $\alpha=0.1$ & $\alpha=0.3$ & $\alpha=0.5$  & $\alpha=0.7$ & tIoU \\ \hline
          \hline
            MCN  \cite{Hendricks_2017_ICCV} & 42.80 & 21.37 & 9.58  & - & 15.83  \\
            CTRL \cite{Gao_2017_ICCV} & 49.09 & 28.70  & 14.00 & - & - \\
            ACRN \cite{yuan2018find} & 50.37 & 31.29 & 16.17 & - & - \\
            MLVI \cite{xu2019multilevel} & - & 45.30  & 27.70  & 13.60  & - \\
            TGN  \cite{chen-etal-2018-temporally} & 70.06 & 45.51 & 28.47 & - & - \\
            TripNet \cite{hahn2019tripping} & - & 48.42 & 32.19 & 13.93 & - \\ 
            ABLR  \cite{yuan2018find} & 73.30 & \textbf{55.67} & \textbf{36.79} & - & - \\
            \hline
            Ours & \textbf{75.25} & 51.28 & 33.04 & \textbf{19.26} & - \\ \hline
    \end{tabular}
    }
    \caption{Accuracy on ANet-Cap for different tIoU $\alpha$ levels.}
    \label{tab:anetcap:res}
    \vspace{-2mm}
\end{table}

\begin{figure*}[!ht]
    \centering
    \includegraphics[width=0.96\textwidth]{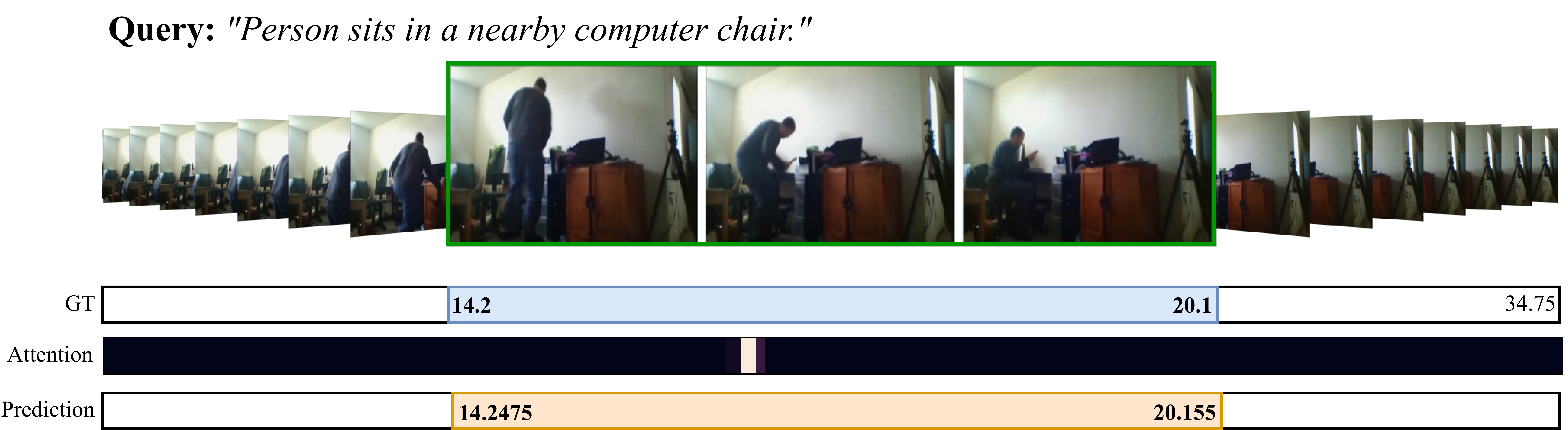}
    ~
    \includegraphics[width=0.96\textwidth]{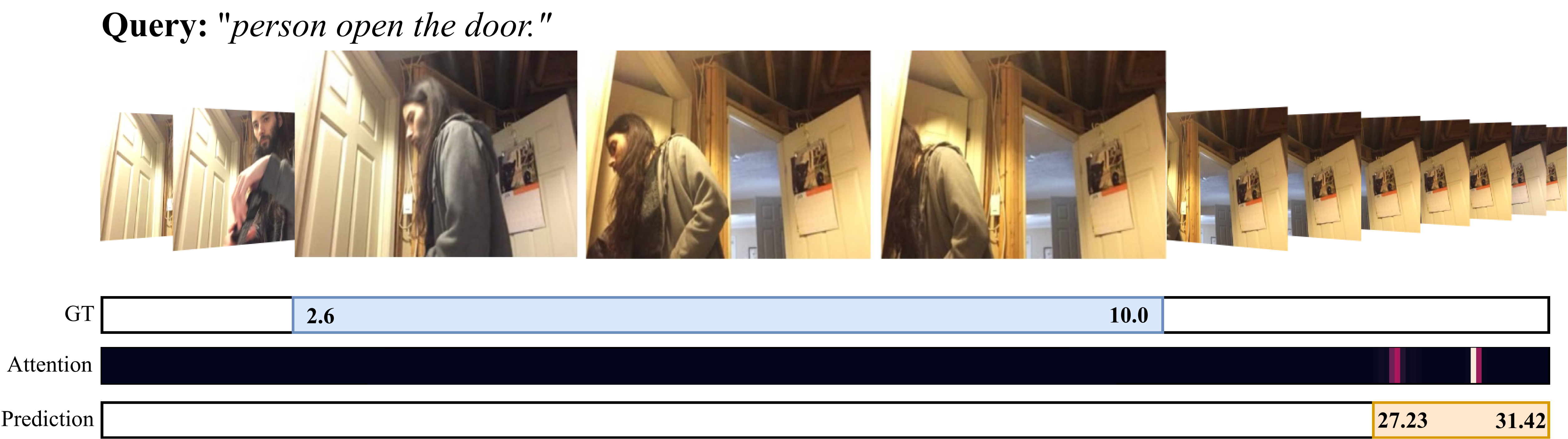}
    \caption{Examples of success and failure cases of our method for Charades-STA.}
    \label{fig:qualitative_charades}
    \vspace{-3mm}
\end{figure*}

As suggested by the empirical evidence, our method consistently outperforms others on estimating the correct extension of the queries. This indicates that our proposed mechanism for incorporating the uncertainty of the labeling process is effective yet simple, playing a key role on helping the network to find the correct starting and ending points. In addition to this, the evidence also suggest that our novel attention mechanism, which guides the localization layer to focus on the features that are within the corresponding segments in the video also aids the network. By allowing the model to attend the features that better represent similar action across different videos, we obtain better generalization.

\begin{table}[]
    \centering
    \scalebox{0.85}{
    \begin{tabular}{|c|c|c|c|c|c|}
        \hline
         \textbf{Method} & MCN  & CTRL & ACRN  & ABLR  & Ours\\ \hline
         \textbf{Mean tIoU} & 15.83 & 20.54 & 24.16 & 36.99 & \textbf{37.78} \\ \hline
    \end{tabular}}
    \caption{Mean tIoU in the ANet-Cap benchmark.}
    \label{tab:anet:miou}
    \vspace{-4mm}
\end{table}

\begin{figure}[t]
    \centering
    \includegraphics[width=0.35\textwidth]{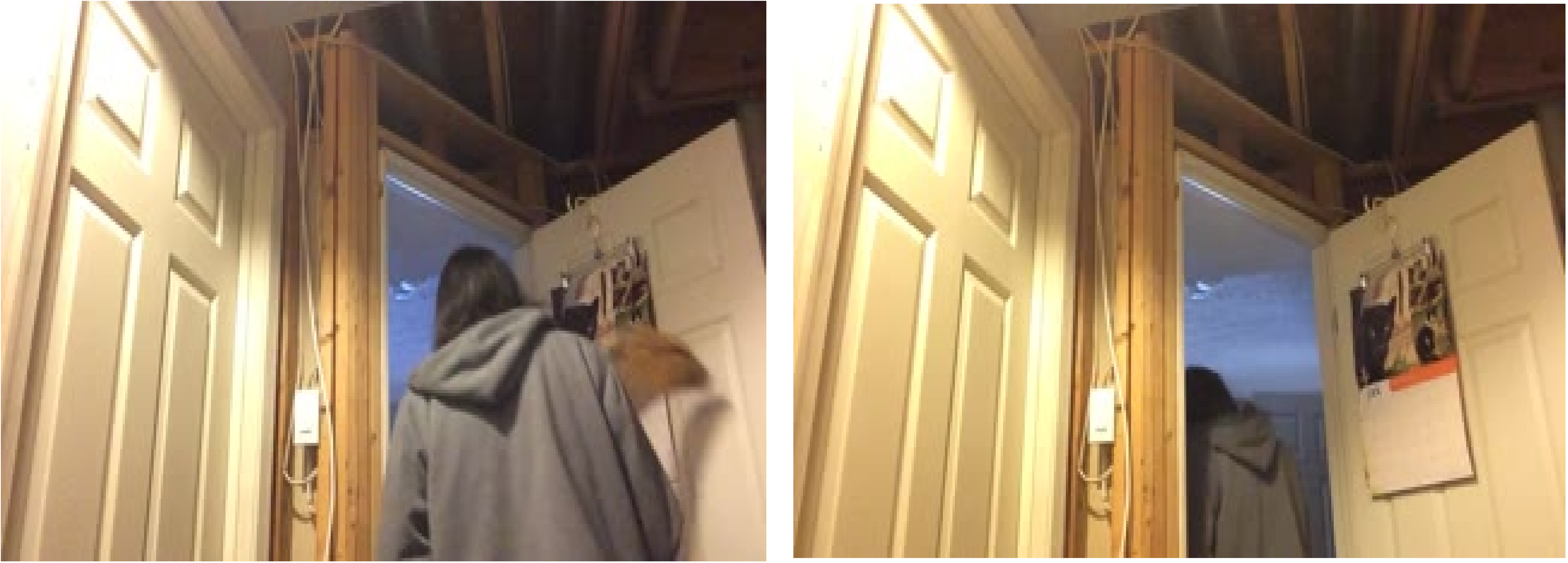}
    \caption{Similar appearance frames for failure case on Charades-STA}
    \label{ambiguity}
    \vspace{-5mm}
\end{figure}
\subsection{Qualitative Results}

Two different samples, one showing a success and one a failure case of our method on Charades-STA dataset can be seen in Figure \ref{fig:qualitative_charades}. Each sample presents the ground truth localization, the attention weights and predicted localization of a given query. For the attention, brighter colors mean more weight. In the successful case, given the query {\em ``Person sits in a nearby computer chair.''} our method could localize the moment at a tIoU of 98.28\%, with a maximum attention at 16.27 seconds peaking at 0.83. It is interesting to see that only one or two video features seem to be necessary for retrieving the starting and ending correctly.

On the second example in Figure \ref{fig:qualitative_charades} we show how our method fails to localize the query {\em ``person open the door''}. It is possible to see that the appearance of the retrieved moment, when the person actually leaves the room, is very similar to the ground truth, Figure \ref{ambiguity}. We attribute this result to the features for opening the door and leaving the room being too close, especially on this example. Probably high quality spatio-temporal features or deeper reasoning about the context would help the network to disambiguate this type of scenarios. More qualitative experiments of success and failure cases of our method on Charades-STA and ANet-Cap can be find on the supplementary material.


%% file: conclusion.tex
\section{Conclusion}
In this paper we have presented a novel end-to-end architecture that is designed to address the problem of temporal localization of natural-language queries in videos. Our approach uses a guided attention mechanism that focus on more generalizable features to guide the localization estimation. Moreover, we also consider the key problem of subjectivity in the annotation process by modeling the label uncertainty in a simple but efficient way, also obtaining substantial performance gains. As a result, our approach archives state-of-the-art performance on three challenging datasets.

